\documentclass{optica-article}

\journal{opticajournal} 

\articletype{Research Article}

\usepackage{lineno}
\usepackage{subcaption}
\usepackage[ruled,vlined]{algorithm2e}
\usepackage{multirow,booktabs}
\usepackage{makecell}
\usepackage{cancel}

\usepackage{soul}
\usepackage{siunitx}

\usepackage{color}

\begin{document}

\title{Improving Fast Auto-Focus with Event Polarity}

\author{Yuhan Bao,\authormark{1,2} Lei Sun,\authormark{1,2} Yuqin Ma,\authormark{1,2} Diyang Gu,\authormark{1,2} and Kaiwei Wang\authormark{1,2,*}}

\address{\authormark{1}State Key Laboratory of Modern Optical Instrumentation, Zhejiang University, Hangzhou 310027, China\\
\authormark{2}National Engineering Research Center of Optical Instrumentation, Zhejiang University, Hangzhou 310027, China\\}

\email{\authormark{*}wangkaiwei@zju.edu.cn} 

\begin{abstract*} 

Fast and accurate auto-focus in adverse conditions remains an arduous task. The emergence of event cameras has opened up new possibilities for addressing the challenge. This paper presents a new high-speed and accurate event-based focusing algorithm.
Specifically, the symmetrical relationship between the event polarities in focusing is investigated, and the event-based focus evaluation function is proposed based on the principles of the event cameras and the imaging model in the focusing process.
Comprehensive experiments on the public event-based autofocus dataset (EAD) show the robustness of the model. Furthermore, precise focus with less than one depth of focus is achieved within 0.004 seconds on our self-built high-speed focusing platform. The dataset and code will be made publicly available. 
\end{abstract*}

\section{Introduction}

Focusing is an indispensable part of any camera. Conventional frame-based camera auto-focus (AF) relies on sharpness analysis of images within a certain exposure time to get the optimal focus position, which makes it difficult to complete focusing in extremely dark and high-speed motion scenes. In recent years, the event camera or Dynamic Vision Sensor (DVS) shows superior performance in temporal resolution (>500Hz) and dynamic resolution (>130dB). Its unique asynchronous event output mechanism allows for a very high response time (order of \si{\micro\second}), meanwhile greatly reducing the bandwidth requirements (compared to the image matrix output of frame-based cameras). Therefore, it has been widely used in object tracking~\cite{RN43, RN44}, Simultaneous Localization and Mapping (SLAM)~\cite{RN46}, visual scene image reconstruction~\cite{RN47}, image motion deblurring~\cite{sun2022event,sun2023event}, human pose estimation~\cite{chen2022efficient}, and many other visual applications.
Likewise, all the advantages of event cameras show their potential in fast AF under adverse conditions like dark, fast-moving, and high dynamic range scenes.
However, event camera AF remains an issue for the reason that the criteria for determining whether an event camera is in focus based on its output information have not been well investigated.

The unique asynchronous output of event cameras makes it impossible to rely on the contrast~\cite{RN48}, statistics~\cite{RN50}, or frequency~\cite{RN49}-based algorithms of conventional frame-based cameras for focusing. Recently, event cameras with integrated conventional frame-based cameras (Active Pixel Sensor, APS) come out, namely the ``Dynamic and Active-pixel Vision Sensor'' (DAVIS)~\cite{RN7, RN8} series from IniVation, which means users can determine whether the event camera is in focus or not by the synchronized frame output. However, the limited spatial resolution and relatively high-noise characteristics of the APS, make auto-focus challenging. In addition, such a focusing method does not take advantage of the high temporal resolution characteristic of event cameras, making the focusing process as slow as with conventional frame-based cameras. Another event camera focusing idea is to use the output of the event by the event camera over a period of time to accumulate into event frames and analyze their contrast, which is a recommended way for DAVIS focusing~\cite{RN52}. However, this is quite time-consuming and requires a specific focus board (e.g., Siemens Star) to complete a focus session, resulting in restricted application scenarios. Recently, some event-rate-based methods have provided new ideas for event camera AF~\cite{RN53}. Previous event-based AF algorithms obtain the focus position by finding the position with the highest event rate (ER) in the whole focus sequence, which sometimes leads to focus errors of 5 to 6 depths of focus. Such an error comes mainly from three reasons: (1) positive and negative event polarity rate peaks separation, (2) the imbalance of positive and negative event polarity ratio, and (3) the strong coupling relationship between ER and scene optical flow. Such a large focusing error is unacceptable for vision tasks, so a faster and more accurate event camera AF algorithm needs to be developed.
\begin{figure}[htbp]
    \centering
    \vspace{-5pt}
    \includegraphics[width=11cm]{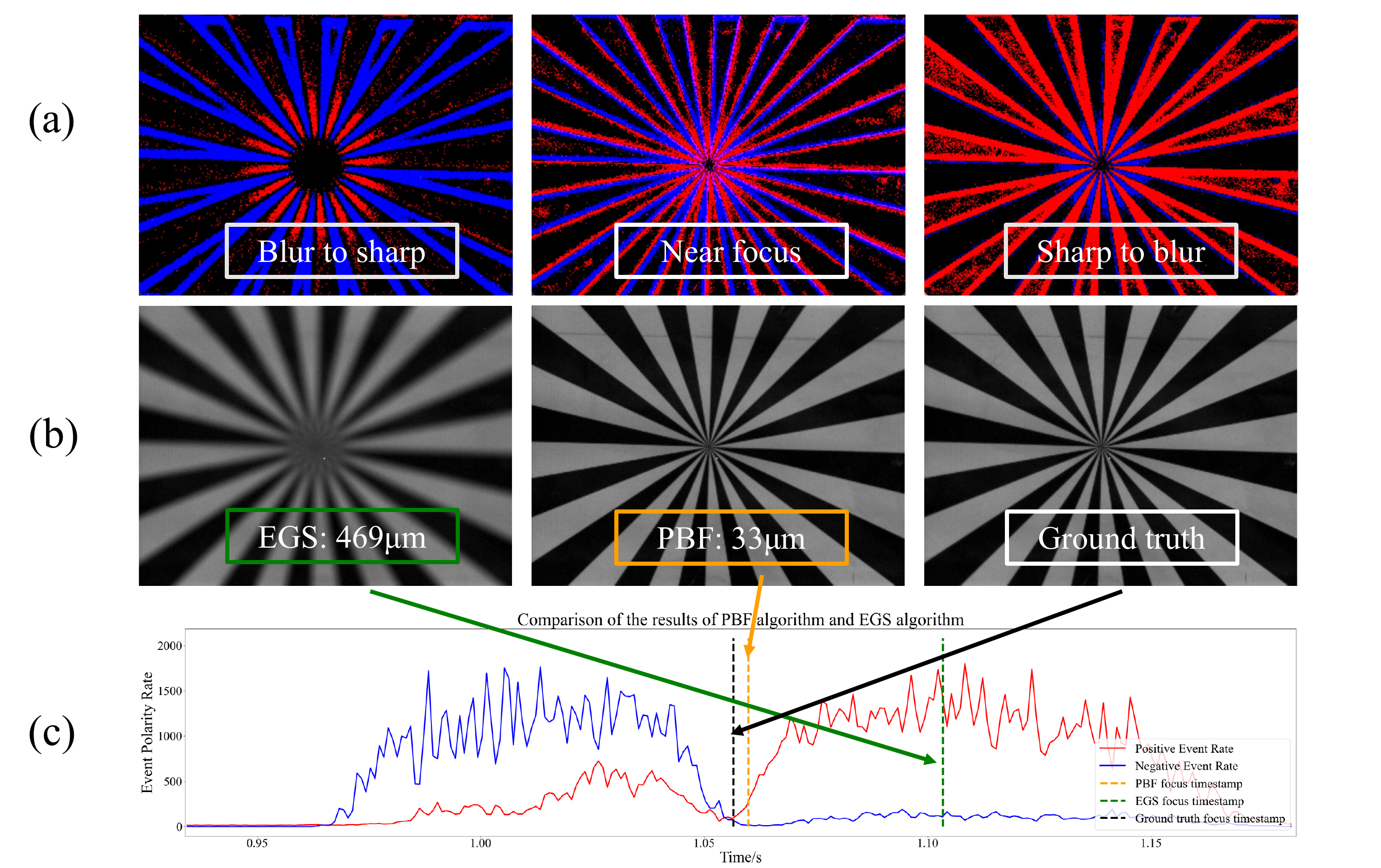}
    \vspace{-7pt}
    \caption{
    Focus result comparision of proposed PBF method and EGS method in a given scenario. (a) Event polarity cumulative images of blurry to sharp (left) and sharp to blurry (right) and near focus (middle). Blue dots: negative events, red dots: positive events. (b): The frames corresponding to the positions provided by EGS algorithm, PBF algorithm and ground truth. (c): The event polarity rate curves during the focusing motion.}\label{fig:Intro}

\end{figure}
During the focusing process, the brightness changes from blurry to sharp and from sharp to blurry are opposite for the same feature point. Since event cameras can respond to changes in brightness in the scene (darkening triggers negative events and brightening triggers positive events), the polarity of the events generated by the same feature should be opposite before and after the focus position, as shown in Fig.~\ref{fig:Intro} (a). Based on such a polarity symmetry phenomenon, the events of different polarities are stacked into two event polarity rate (EPR) sequences and a focusing evaluation function is constructed to evaluate the symmetry of the sequences. Fig.~\ref{fig:Intro} (c) shows the focus result of (a) given by our proposed ``Polarity-based auto-focus'' (PBF) method, leveraging the symmetry of EPR sequences, as well as the state-of-art algorithm ``event-based golden search'' (EGS)~\cite{RN53}, leveraging the highest ER. Fig.~\ref{fig:Intro} (b) intuitively shows the frames corresponding to the positions provided by EGS algorithm, PBF algorithm and ground truth.


In summary, the four contributions of this article are listed below:
\begin{itemize}
    \item From the imaging principle of the focusing process and the mechanism of event camera signal generation, the properties of the EPR variations in the focusing process are well elucidated.
    \item A universal event camera auto-focusing algorithm, polarity-based auto-focus (PBF), is derived from the correlation between focus position and EPR, which can be applied to most event camera devices. And the codes are available in Code 1 (Ref.~\cite{PBFcode}).
    
    \item Extensive experiments are conducted on the public ``event-based auto-focus dataset'' (EAD)~\cite{RN54}, as shown in Dataset 1 (Ref.~\cite{EADnpy}), which includes challenging scenarios such as low light, rapid motion, and high-speed changes. The proposed PBF outperforms the state-of-the-art EGS algorithm by 17.5\% in RMSE accuracy and 185 times faster on average in processing speed.

    \item An event camera high-speed focusing (EvHF) dataset featuring scenes with varying lighting conditions and motion states is assembled, as shown in Dataset 2 (Ref.~\cite{highspeeddataset}), on which our proposed PBF achieved a focusing error of less than 1 depth of focus with a processing time of fewer than 0.004 s, which outperforms EGS by 5 times in MAE accuracy and 36 times in average speed.
    
\end{itemize}
The rest of the article is organized as follows. Section~\ref{section:2} presents related works, including the state-of-art algorithm EGS. In Section~\ref{section:3}, the sources of event camera signals during the focusing process and the intrinsic connection between the focus position and EPR are analyzed in detail. Then, the processing of the EPR sequences and  our PBF algorithm are introduced. Section~\ref{section:4} shows the results of our proposed algorithm on the simulation dataset, the public dataset EAD and our high-speed focusing dataset EvHF, as well as special scenarios where brightness changes abruptly during focus. At the end of this section, ablation experiments are performed to demonstrate the necessity of the method mentioned in our algorithm. Finally, the conclusion and discussion of this paper are provided in the rest sections.

\section{Related works}
\label{section:2}
\subsection{Conventional AF methods}
The convenience of digital image acquisition makes most of the current conventional camera AF algorithms use methods that analyze image information, such as the contrast analysis method~\cite{mir2014extensive} and phase analysis method~\cite{chan2018improving}. However, dynamic and extremely dark scenes pose a great challenge to the above frame-based methods because of the need for a longer exposure time to obtain imaging signals. Visualization 1 in Ref.~\cite{repeated_focus} shows a repeated focusing process using a conventional autofocus camera in dark scenes. 
\subsection{Event-based AF}
\begin{figure}[htbp]
    \centering
    \includegraphics[width=12cm]{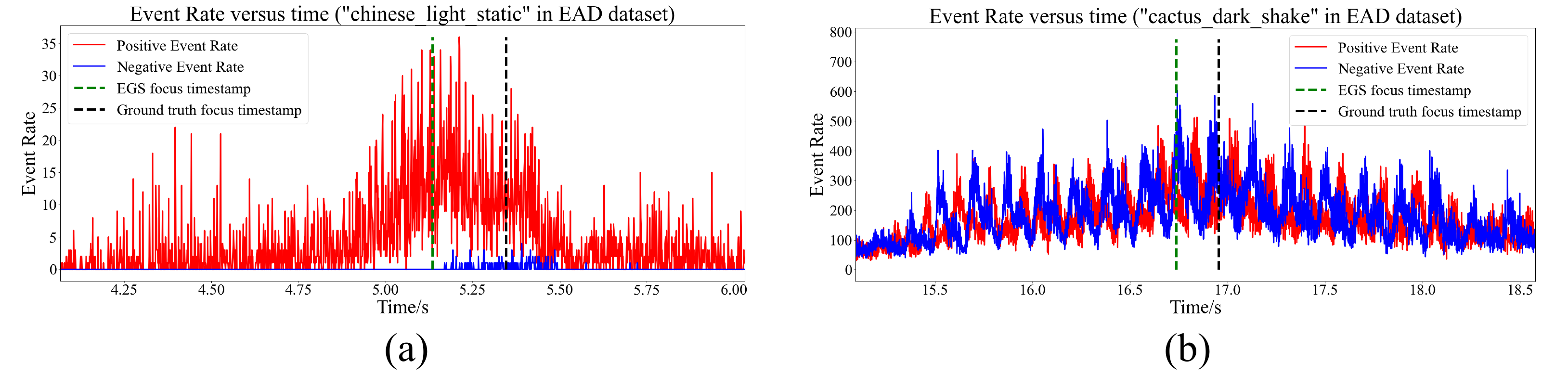}
    \vspace{-5pt}
    \caption{(a) and (b) demonstrate the focus result using EGS algorithm~\cite{RN53}. Green dashed line: the focus timestamp predicted by the EGS. Black dashed line: ground truth. (a): ``chinese light static'' in EAD dataset, which shows an imbalance in EPR sequences. (b): ``cactus dark shake'' in EAD dataset, which shows the impact of violent motion.}
    \label{fig:related works}
\end{figure}

In the past, the approach to AF based on event information has often been to integrate events in time and analyze event accumulated frames, which is quite similar to conventional frame-based AF ways. Recently, Lin et al.~\cite{RN53} found the relationship between ER and focus evaluation function. According to event cameras’ working principle, event cameras are inherently more responsive to edges in the scene with high contrast or image gradients while the traditional frame-based AF methods also leverage the high contrast properties of images in focus. For the entire focusing process, the contrast on the image plane is greatest near the focus position, as shown in Fig.~\ref{fig:Intro} (b), and thus the event camera outputs the most events as well. Based on the above ideas, they proposed using ER as an event-based focus measure. They used a dichotomous lookup algorithm called Event-based Golden Search (EGS) to find the time point with the highest ER during the whole focusing process. Their experimental results show that the event-based focusing algorithm outperforms the conventional frame-based focusing algorithm in dark and dynamic scenes. However, their algorithm does not meet the real-time requirements in many cases. Also, their algorithm is not robust to event polarity imbalance, with -191.0 \si{\micro\meter} error, approximately 5 depths of focus in Fig.~\ref{fig:related works} (a), and violent motion, with -236 \si{\micro\meter} error, approximately 6 depths of focus in Fig.~\ref{fig:related works} (b).


According to Lin's derivation, the ER is proportional to the product of the image gradient and the image optical flow. The image gradient $\nabla I(\boldsymbol{x},t)$ is the term closely related to the degree of focus, and using the ER as the focus evaluation function requires the optical flow term $\boldsymbol{v}(\boldsymbol{x},t)$ to be constant, which is usually difficult to achieve.
\begin{equation}
    \label{eq1}
    ER(\boldsymbol{x},t,\Delta t) =\frac{\nabla I(\boldsymbol{x},t)}{C}\cdot \boldsymbol{v}(\boldsymbol{x},t),
\end{equation}

where the boldsymbol $\boldsymbol{x}$ indicates the pixel index, $t$ stands for the timestamp of inspection, $\Delta t$ is the time interval for calculating ER and $C$ is a constant scalar.

The work of Ge et al.~\cite{RN55}. is closest to our idea, which applies the event camera in AF for small depth-of-focus microscopes. Interestingly, they found that events were mainly triggered symmetrically at both sides near the focal plane, the event polarities were different between the two sides and events were almost undetectable at the focus position. This appears to be the exact opposite of the EGS algorithm idea introduced above, which regards the position with the highest ER as the focus position. In reality, however, such opposing views are both reasonable, given the differences in their experimental conditions. Ge et al. focus on the static scene without optical flow, while Lin et al. focus on more general situations, such as handheld shooting and scenes in constant optical flow.

Both of their schemes get relatively good results in constrained scenes, while a universal event camera AF algorithm does not want to be constrained by any scene limitations (e.g., stationary or uniform motion). In the next section, the principle of event generation during focusing is elaborated in detail to give a universal event camera AF model.

\section{Proposed method}
\label{section:3}
\subsection{Principle of event camera}
\label{subsection:31}

\begin{figure}[ht!]
\vspace{-5pt}
\centering\includegraphics[width=9cm]{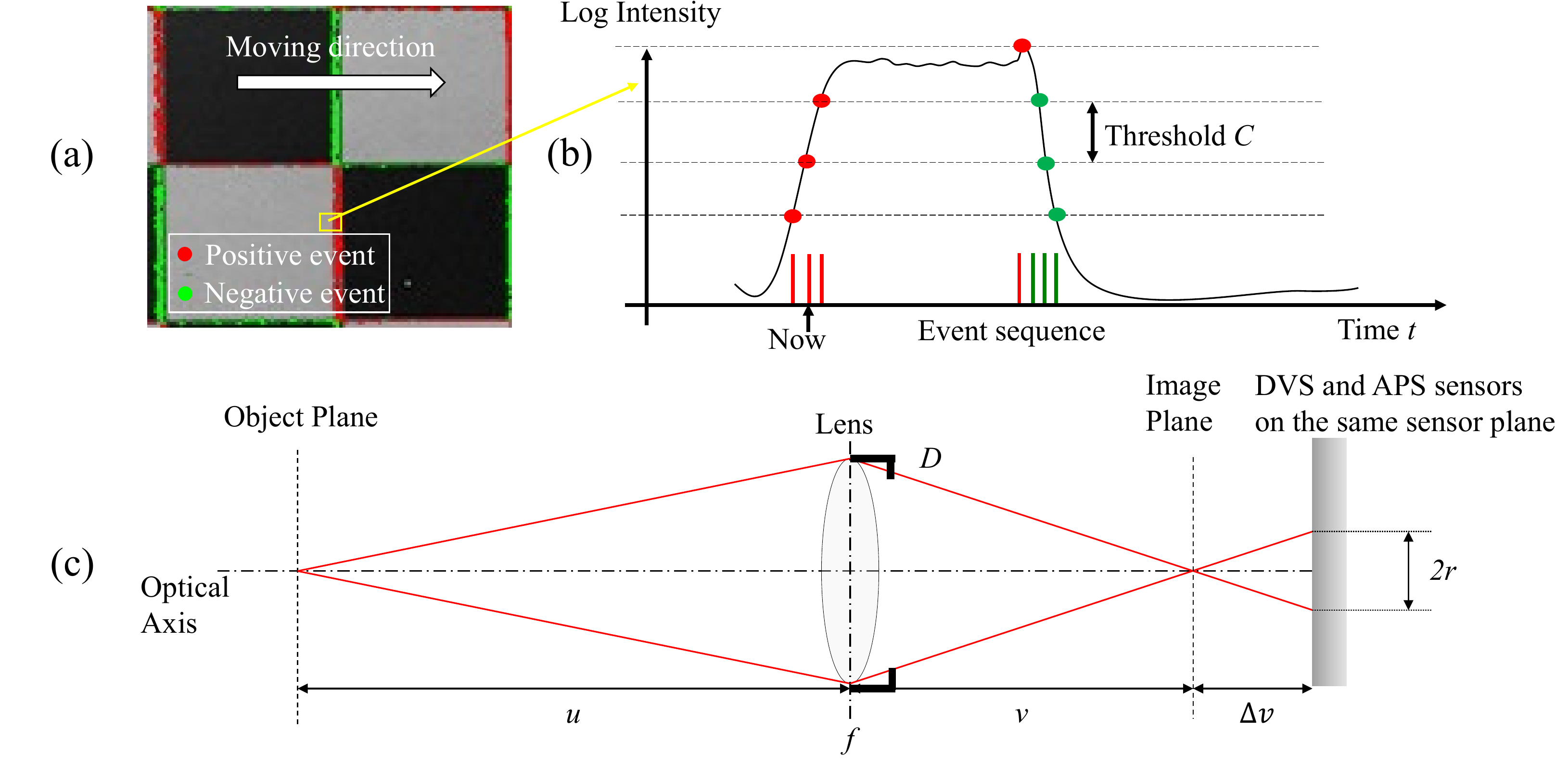}
\vspace{-7pt}
\caption{Event camera principle and optical principle. (a) and (b) illustrate the working principle of the event camera. The checkerboard grid in (a) moves from left to right, triggering events on a pixel (yellow square) at the edge. (b) shows the brightness change of the pixel and the corresponding event sequence. (c) shows the ideal imaging model of a thin lens in an event camera. DVS and APS sensors share the same sensor plane.}
\vspace{-8pt}
\label{fig: Priciple}
\end{figure}

Event cameras respond to brightness changes and output events in an asynchronous manner. For each pixel, once the log of intensity $L(x,y,t)$ changes above a set threshold $C$, it emits an event signal and assigns positive or negative polarity $p$.

\begin{equation}
\label{eq2}
p{(x,y,t_i)}=\left\{
\begin{aligned}
+1 & , & L(x,y,t_i)-L(x,y,t_{i-1})>C, \\
-1 & , & L(x,y,t_i)-L(x,y,t_{i-1})<-C,
\end{aligned}
\right.
\end{equation}
 where $t_i$ indicates the current timestamp, while $t_{i-1}$ indicates the last timestamp when an event is triggered.
As shown in Fig.~{\ref{fig: Priciple}} (a), the projection of a rigid edge moves from left to right, and the pixels near the edge trigger events of different polarities (green for negative, red for positive). (b) shows the brightness change of the pixel during the movement. At this time point, the brightness is raised above the threshold due to the white checkerboard on the left side passing by the pixel marked with a yellow box, triggering a series of positive red events; after some time, the next edge passes by the pixel and the brightness decreases, triggering a number of green negative events.

Due to device manufacturing issues, the actual threshold value for each pixel triggering event is not necessarily the set value $C$. Even the thresholds for a specific pixel triggering positive and negative events are not always the same, thus resulting in the polarity imbalance phenomenon shown in Fig.~\ref{fig:related works} (a). 

Differences in positive and negative event thresholds often pose significant problems in the task of visual reconstruction of event information~\cite{RN57}. In fact, polarity threshold compensation~\cite{RN56} should be considered in any application that considers event polarity. In the post-processing of event data, the problem of different positive and negative thresholds can be compensated by normalizing the sequence of positive and negative event rates. 

In the next subsection, the sources of events generated by the event camera during focusing are described.
\subsection{Source of events during focusing}
\label{subsection:32}
In general, events arise from changes in the brightness of the sensor plane. More precisely, in the scene of AF, the brightness change comes from 4 sources: (1) changes in the convolutional blur kernel during focusing, (2) focus breathing, (3) optical flow caused by motion, and (4) unexpected noise due to devices. In the following four subsections, each of these four sources is elaborated on.
\subsubsection{Blur kernel changing during focusing}
\label{subsubsection:321}
Fig.~\ref{fig: Priciple} (c) shows the ideal imaging
model of a thin lens and the ideal imaging equation is as follows:
\begin{equation}
\label{eq4}
\frac{1}{u}+\frac{1}{v}=\frac{1}{f},
\end{equation}
where $u$ represents the object distance, $f$ represents the focal length of the thin lens, and $v$ represents the ideal image distance.

The event camera and frame-based sensors share the same sensor plane, so the event camera's focus judgment is consistent with the frame-based camera. Due to the gap between the sensor plane and the image plane, the ideal image point on the image plane is blurry on the sensor plane, and the radius of the dispersion circle is $r$. According to the simple geometric similarity relationship, the radius of the dispersion circle $r$ can be derived:
\begin{equation}
\label{eq5}
r={\frac{\left| \Delta v \right|}{2v_0}}D,
\end{equation}
where $\Delta v$ , $v_0$ , and $D$ represents the defocus amount, the ideal image distance, and the exit pupil of the lens, respectively.

In the case of incoherent light illumination, the point spread function of a circular aperture is the square of a first-order Bessel function. Due to the complexity of the Bessel function, a two-dimensional Gaussian function is often used for approximation in practical applications:
\begin{equation}
\label{eq6}
h(x,y,a)=\frac{1}{2\pi a^2}\exp{({-\frac{x^2+y^2}{2a^2}})},a=kr,
\end{equation}
where $k$ equals to the inverse of the sensor pixel size, while $(x,y)$  is the coordinate of the pixel.

\begin{figure}[ht!]
 \vspace{-5pt}
\centering\includegraphics[width=10cm]{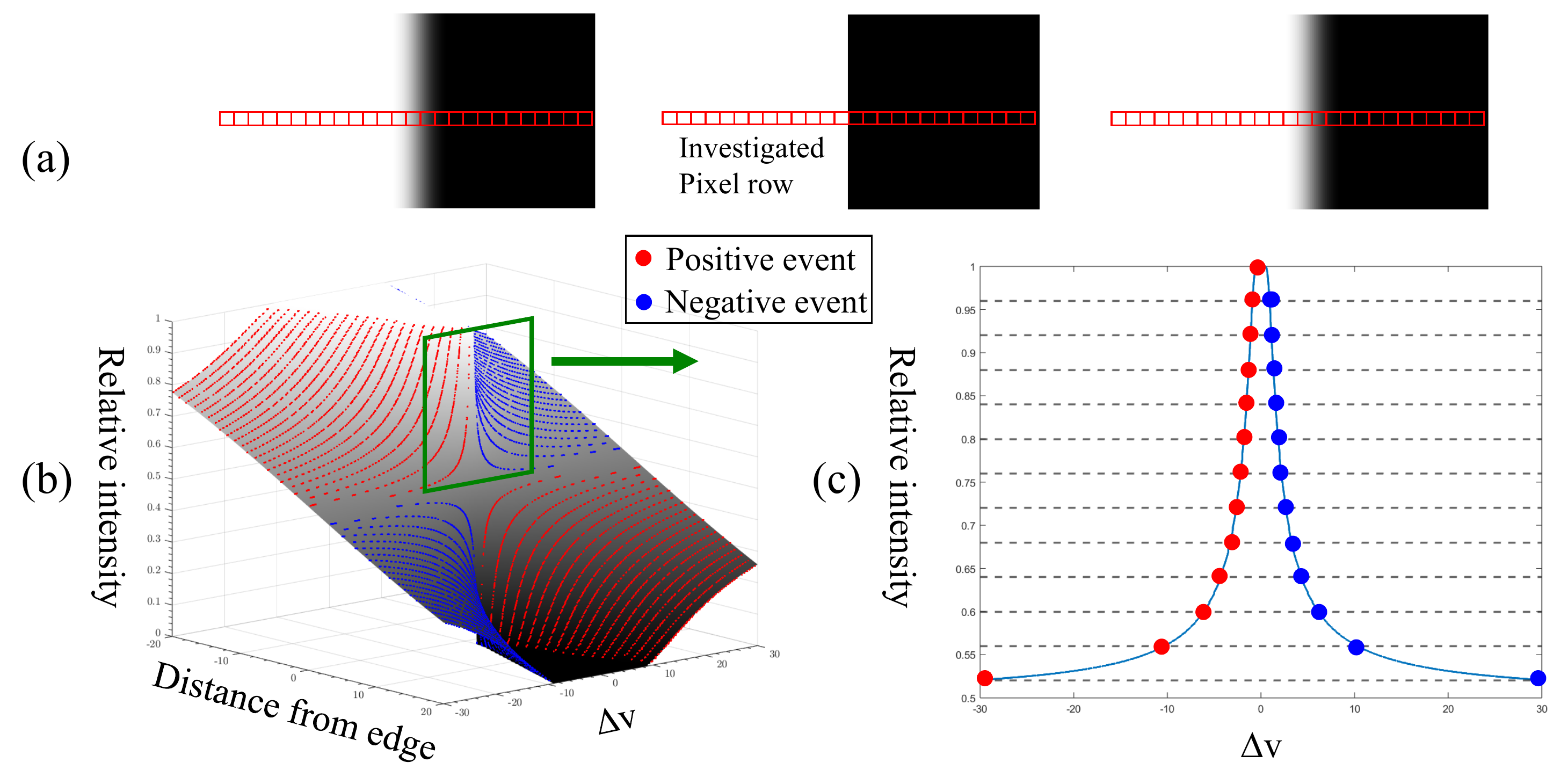}
 \vspace{-5pt}
\caption{Change of log light intensity near the edge during focusing and triggered events. (a): The edge feature changes during focusing. (b): The change of relative log intensity of the investigated pixel row, where the red and blue dots indicate the positive and negative events triggered in the process. (c): The relative log intensity change of the pixel closest to the edge, and the corresponding event.
 }\label{fig: blur kernel changing}
 \vspace{-8pt}
\end{figure}

Fig.~\ref{fig: blur kernel changing} (a) shows a sharp edge changing during focusing. (b) shows the change in brightness and the triggered events of the investigated pixel row when $\Delta v$ changes from -30 to +30. (c) shows the relative brightness change of the pixel closest to the edge. In the ideal case, the polarity of events triggered by a single pixel before and after the focus position exhibits perfect symmetry.

When considering a large region of interest (ROI), the effect of log intensity changes of all pixels within the ROI needs to be accumulated. From the derivation of Eq.~(\ref{eq5}) and Eq.~(\ref{eq6}), the convolution kernel size $a$ is proportional to the absolute value of the defocus amount $\Delta v$. The ideal image distance is $v_0$, then the defocus amount $\Delta v= v-v_0$. The positive event rate is denoted as PER and the negative event rate as NER. Then Eqs.~(\ref{7a}), (\ref{7b}) and (\ref{7c}) are derived.

\begin{subequations}
\label{eq7}
\begin{align}
PER(v)=\sum_{\frac{\partial}{\partial v}L(x_i,y_i,v)>0,(x_i,y_i)\in ROI}\frac{\partial}{\partial v}L(x_i,y_i,v), \label{7a}\\
NER(v)=\sum_{\frac{\partial}{\partial v}L(x_i,y_i,v)<0,(x_i,y_i)\in ROI}\frac{\partial}{\partial v}L(x_i,y_i,v),  \label{7b} \\
L(x_i,y_i,v)=\log(I\otimes h(x_i,y_i,k\frac{\left|v-v_0\right|}{2v_0}D), \label{7c}
\end{align}
\end{subequations}

where the variable $I$ refers to the light intensity matrix on the sensor plane, and the $\otimes$ symbol indicates the convolution of $I$ with the point spread function $h$ at pixel $(x_i,y_i)$ when the image distance is $v$.

\begin{figure}[ht!]
\centering\includegraphics[width=11cm]{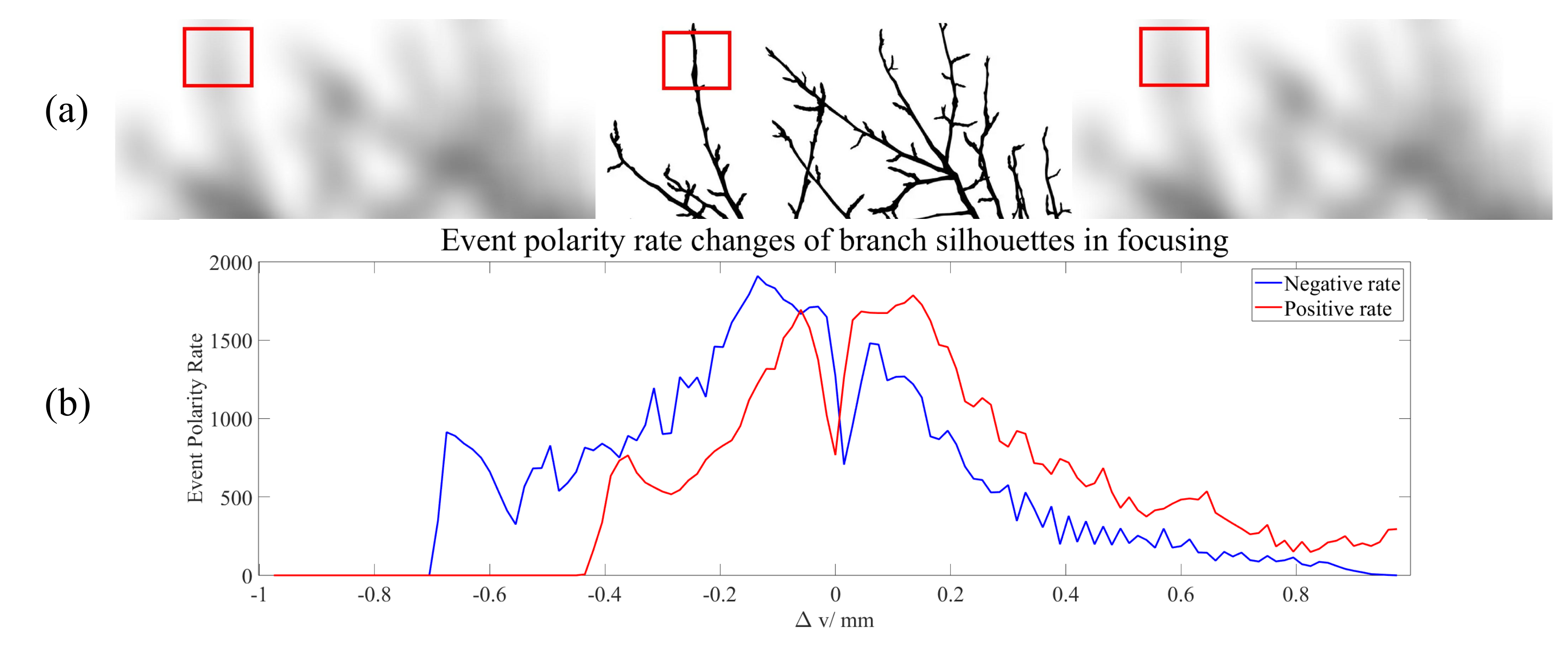}
\caption{EPR changes of branch silhouettes in focusing. (a): The frame image of branch silhouettes during focusing. (b) The curve of EPR in the red box (ROI) during focusing, which shows symmetry.
 }\label{fig:branches example}
 \end{figure}

According to Eq.~(\ref{7c}), the log intensity values $L(x_i,y_i,v)$ of a given pixel are equal when $v$ is symmetric about $v_0$. After accumulating the intensity changes for all pixels, the PER sequence and the NER sequence should also be symmetrically distributed about $v_0$. The simulated example in Fig.~\ref{fig:branches example} provides a more intuitive view of the symmetry relationship of the EPR, where (a) shows the process of focusing on a branch, and the red box indicates the ROI. Since the differential is replaced with a difference in the $v$-direction here, and the threshold is not infinitesimal, the rounding error leads to an imperfect symmetric curve in Fig.~\ref{fig:branches example} (b). 

 To summarize this subsection, (1) the image change during the focusing process can be described by the change in the size of the Gaussian blur kernel. (2) The size of the Gaussian blur kernel is proportional to the defocus amount $\Delta v$. (3) The EPR of a pixel is proportional to the log intensity change rate of that pixel. And (4) the log intensity changing is symmetrically about the focus position $v_0$, and subsequently, the PER and NER curve should be symmetrically about $v_0$.

\subsubsection{Focus breathing}
The field of view (FOV) of a lens slightly changes during focusing~\cite{RN61}, which is called focus breathing. It introduces additional optical flow, which points to the center of the image due to the axial symmetry of the optical system. Such changes are nearly linearly related to the defocus amount~\cite{Goodsell:22}. Consequently, when the image plane moves at a uniform speed, the resulting optical flow is also constant. According to Eq.~(\ref{eq1}), the events caused by the breathing effect peak when the image gradient is at its maximum, i.e., on focus. In fact, events generated by the breathing effect are necessary for the EGS algorithm in static scenes because of the necessary constant optical flow $\boldsymbol{v}(\boldsymbol{x})$.

For our method, the events brought about by this constant optical flow do not affect the symmetry of the PER and NER, because these events remain symmetric about the focus position, referring to Eq.~(\ref{eq1}) and Eq.~(\ref{eq7}). 

Actually, the main problem caused by the breathing effect is that the same feature is located at different pixels at different image distances. When a larger ROI is considered for focusing, since the entire ROI-generating events are accumulated as a whole as illustrated in Eq.~(\ref{eq7}), the feature movement due to the breathing effect does not interfere with the focusing task. The breathing effect only needs to be considered when reaching a pixel-level focus judgment. 

\subsubsection{Optical flow caused by motion}
\label{subsubsection:323}
The events caused by dynamic scenes and camera motion are the biggest interference in event camera AF. In a dynamic scene (such as ``cactus dark shake'' in the EAD dataset shown in Fig.~\ref{fig:related works} (b)), the ER changes dramatically due to the drastic round trip camera movement, which means that the EGS algorithm is not robust to these irregular jitters, since Eq.~(\ref{eq1}) requires a constant optical flow $\boldsymbol{v}(\boldsymbol{x})$.
Eq.~(\ref{eq1}) and Eq.~(\ref{eq7}) show the two main sources of events in the focusing process: optical flow due to motion in the scene (from both camera and object), and the brightness changes due to focusing. The change in the defocus amount $\Delta v$ can also be considered as a motion as well as the optical flow. The optical flow is perpendicular to the optical axis while the focusing is along the optical axis. 

In the real scene, when the focusing motion is fast enough, such as with the ultrasonic motors widely used in digital cameras, the events generated by optical flow are much fewer than those generated by focusing, which means Eq.~(\ref{eq7}) better describes the output characteristics of events in high-speed focusing scenarios. In fact, the application of event cameras in AF is precisely to pursue ultra-fast focusing, where the optical flow motion can be seen as stationary during the focusing done in a very short time (e.g., several milliseconds). Therefore, the events brought by the optical flow are considered as noise in our method. This event noise constitutes a random high-frequency part of the overall EPR sequence, and it can be filtered out by certain filtering methods.

\subsubsection{Noise}
\label{subsubsection:324}
On event cameras with integrated APS, bursts of DVS events can be caused by the capture of an APS frame. These event burst spikes are conspicuous in Fig.~\ref{fig:related works} (a). In addition, in extremely dark environments, the event camera exhibits significant dark noise.
Due to the noise, it's impossible to just use the ER algorithm in ~\cite{RN53} or the ``M-shaped valley'' algorithm in ~\cite{RN55}. 

Different types of noise have different characteristics in the time and frequency domains. The dark noise is prevalent throughout the EPR sequence and has a relatively fixed high-frequency spectral component; the event surge brought about by the APS frame output is related to the APS frame rate, generally 40Hz to 50Hz; the event noise brought about by scene motion or irregular jitter varies in both the time and frequency domains.

\subsection{Polarity-based auto-focus algorithm}
\label{subsction:33}
\subsubsection{EPR sequence preprocess}
\label{subsubsction:331}

Referring to Eq.~(\ref{eq7}), since the image distance $v$ can be regarded as varying uniformly during the focusing process, the events generated by the focusing motion occupy the low-frequency part of the PER and NER sequences. And through the discussion in Section~\ref{subsubsection:324}, the noise is usually distributed in the high-frequency part of the sequence, and its spectrum is changing in the time domain.

Compared with the traditional Fourier transform, the wavelet transform combines the characteristics of frequency domain and time domain analysis and fits scenarios where the signal spectrum varies in the time domain. Due to the complex time-frequency domain properties of EPR sequences, wavelet transform is more suitable for the EPR sequence preprocess.

\begin{figure}[ht!]
\vspace{-5pt}
\centering\includegraphics[width=12cm]{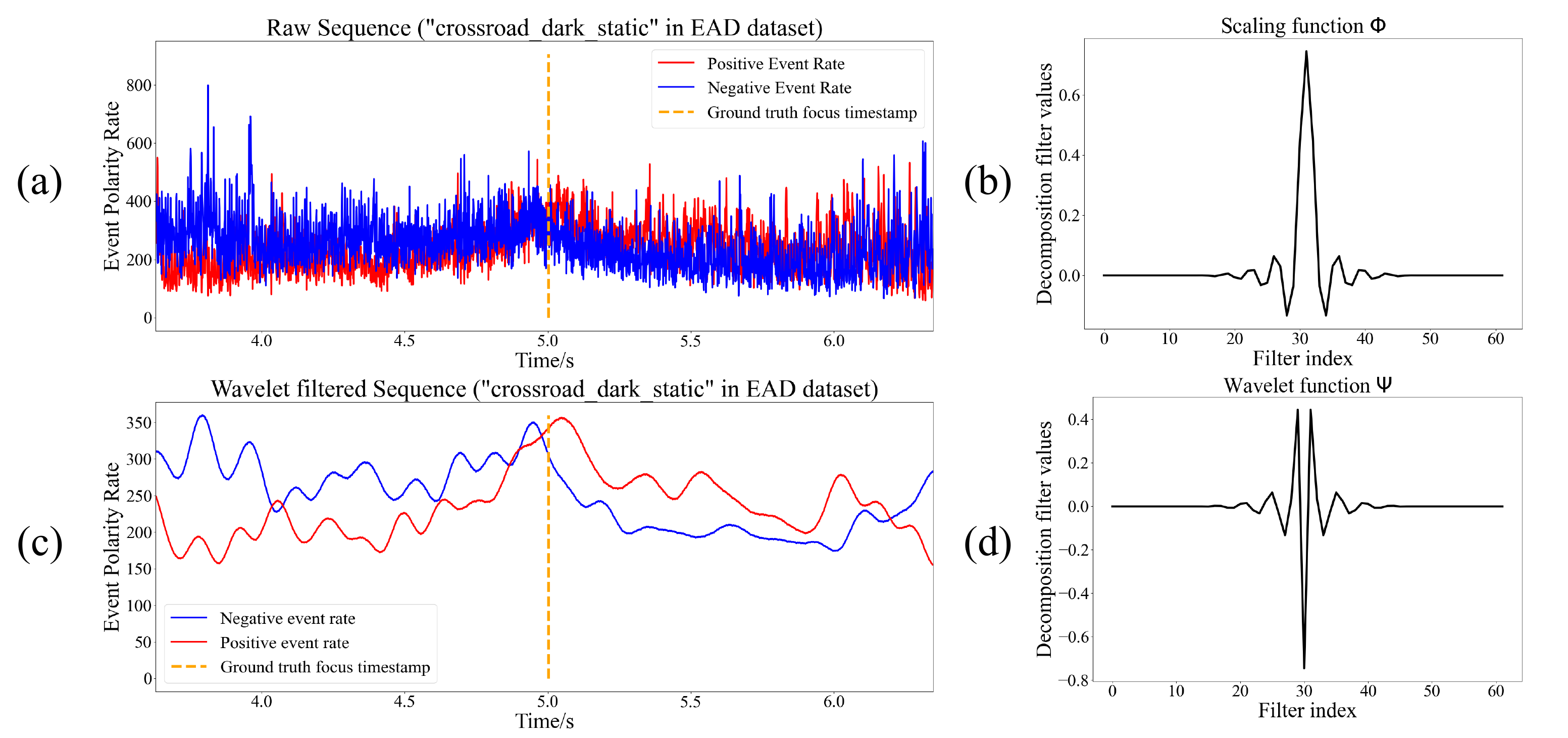}
\caption{Wavelet filtered EPR. (a) EPR of crossroad dark static in EAD dataset, with a lot of noise. (c) EPR after wavelet filtering, showing symmetry. (b) and (d) indicate the scaling function $\phi$ and wavelet function $\psi$ of used wavelet ‘dmey’.}\label{fig: Wavelet result}
\vspace{-5pt}
\end{figure}

The event data in EAD dataset contains a non-negligible amount of the noise described in Sections \ref{subsubsection:323} and \ref{subsubsection:324} because of the slow focusing speed (1$mm$/$s$). As Fig.~\ref{fig: Wavelet result} (a) shows, the curve exhibits violent oscillations because of the noise mentioned in Section~\ref{subsubsection:324}: the dark noise makes up the highest frequency component; the APS noise is the 40Hz fixed interval spikes; and a blue spike noise brought by the scene motion appears at 3.6 s.
The ``discrete Meyer'' (dmey) wavelet shown in Fig.~\ref{fig: Wavelet result} (b) and (d) is used to perform 6-layer wavelet noise reduction on the raw sequence, removing all the high-frequency components and retaining only the last low-frequency component ``cA6''. The filtered sequence is shown in Fig.~\ref{fig: Wavelet result} (c). It can be seen that the low-frequency components of the PER and NER sequences show symmetry consistent with the theoretical analysis in Section \ref{subsubsection:321}.

\subsubsection{Focus evaluation function}
\label{subsubsection:332}
After applying wavelet filtering, the EPR sequence has been effectively purged of noise interference. However, the task of determining the focus position remains. As expounded upon in Section \ref{subsubsection:321}, the EPR distribution is symmetric with respect to the focus position. In order to quantitatively characterize this symmetry and ascertain the focus timestamp of the sequence, a mean square error (MSE) evaluation function similar to the convolution operation is constructed.

Given a denoised PER sequence $P(v)$ and a denoised NER sequence $N(v), v\in(v_1,v_2)$. Firstly, the domain of definition of $v$ is shifted to $(0,v_2-v_1)$,
\begin{subequations}
\label{eq8}
\begin{align}
P'(v)=P(v+v_1),\label{8a}\\
N'(v)=N(v+v_1).\label{8b} 
\end{align}
\end{subequations}
The mean square error (MSE) function is defined as:
\begin{equation}
\label{eq9}
MSE(a)=\left\{
\begin{aligned}
\frac{1}{a}\int_{0}^{a}(N'(v)-P'(a-v))^2dv&,&0<a\leq v_2-v_1, \\
\frac{1}{2(v_2-v_1)-a}\int_{a-v_2+v_1}^{v_2-v_1}(N'(v)-P'(a-v))^2dv&,&v_2-v_1<a<2(v_2-v_1).
\end{aligned}
\right.
\end{equation}
The argument of minimal MSE function is related to the focus position:
\begin{equation}
\label{eq10}
    v^*=\frac{1}{2}\mathop{\arg\min}\limits_{a}(MSE(a))+v_1.
\end{equation}
The PBF MSE function can be decomposed into three distinct steps: (1) reversing the sequence of one polarity, (2) aligning the reversed sequence with the other sequence of different lengths, and (3) computing the MSE between the overlapping regions of the two sequences. The MSE function reaches its minimum value when the reversed sequence is optimally aligned with the other sequence, at which point the parameter $a$ of the MSE function is equal to twice the center of symmetry of the original sequences.

To ensure the convolution MSE function accurately reflects the entire sequence, a hyperparameter investigation factor, $k$, is introduced when the overlap is too small in the PBF algorithm. Additionally, to address the imbalance between positive and negative event amounts in Fig.\ref{fig:related works} (a), the PER and NER sequences are normalized. By discretizing Eqs. (\ref{eq8}), (\ref{eq9}), and (\ref{eq10}), and accounting for the instability caused by short overlap lengths, the final PBF algorithm is derived.

\begin{algorithm}
\DontPrintSemicolon
\KwData{positive event rate sequence $P[v]$, negative event rate sequence $N[v]$,sequence definition domain $v\in(v_1,v_2)$, MSE investigation factor $k$.}
\KwResult{Optimal focus position $v^*$.}
\Begin{
$P[v] \longleftarrow wavelet denoise(P[v])$\;
$N[v] \longleftarrow wavelet denoise(N[v])$\;
$P[v] \longleftarrow normalize(P[v])$\;
$N[v] \longleftarrow normalize(N[v])$\;
$P'[v] \longleftarrow P[v+v_1]$\;
$N'[v] \longleftarrow N[v+v_1]$\;
\For{$a=k(v_2-v_1)$ $\to$ $(2-k)(v_2-v_1)$ }{
\If{$a<v_2-v_1$}{
$MSE[a]=\frac{1}{a}\sum_{1}^{a}(N'[v]-P'[a-v])^2$\;
}
\Else{
$MSE[a]=\frac{1}{2(v_2-v_1)-a}\sum_{a-(v_2-v_1)+1}^{v_2-v_1}(N'[v]-P'[a-v])^2$\;
}
}
$v^*=\frac{1}{2}\mathop{\arg\min}\limits_{a}(MSE[a])+v_1$
}
\caption{Polarity-based auto-focus algorithm (PBF).}
\label{algorithm1}
\end{algorithm}


\section{Experiments}
\label{section:4}
In this section, a simulation experiment is conducted to validate the theory outlined in Section~\ref{section:3}. Subsequently, the PBF algorithm is evaluated on the real-world dataset EAD~\cite{RN54} and compared with the currently state-of-art algorithm EGS~\cite{RN53} in Section~\ref{subsection:42}. Then, high-speed AF is tested on our collected EvHF dataset in Section~\ref{subsection:43}. Due to the sensitivity of the event camera to brightness changes, two scenarios of sudden brightness changes, stroboscopic and arbitrary brightness change, are tested in Section~\ref{subsection:brightness change}. And the performance of PBF and EGS algorithms are compared in these scenarios. Finally, extensive ablation experiments are conducted in Section~\ref{subsec:ablation} to demonstrate the effectiveness of the proposed methods in our algorithm.

\subsection{Simulation experiment}
\label{subsection:41}
The event generation model presented in Section~\ref{subsection:31} and the mechanism of events generated in the focusing process discussed in Section~\ref{subsection:32} are used to develop a MATLAB script, as shown in Code 2 (Ref.~\cite{simulationcode}), that simulates the EPR sequences generated during the actual focusing process. To determine the focus position, our PBF algorithm is applied to the simulated data.
\subsubsection{Simulation experiment setup}
Our simulation model first reads in a light intensity image $I$, then convolves it with Gaussian convolution kernel of different sizes, and finally obtains a series of blurry images under different defocus amounts. According to the definition of event camera brightness, these light intensity images $I$ are log-mapped to the brightness image $L$ and stacked in the order of focusing process (the convolution kernel changes from large to small, and then from small to large). Besides, a new image $L_{old}$ is created to store the last trigger brightness of each pixel, and the initial value is the first one of the blurry image stack. For each pixel in the ROI, when Eq.~(\ref{eq2}) is satisfied, an event signal is triggered, and the brightness of the pixel is saved to $L_{old}$. And if $\Delta L$ is a multiple of the thresholds, the number of events is multiple accordingly, consistent with other event camera simulators~\cite{hu2021v2e}. Then, the number of positive and negative events triggered in the whole ROI during a given time interval $\Delta t$ are accumulated to obtain the EPR time sequences.
\subsubsection{Qualitative Analysis}
\begin{figure}[ht!]
\centering\includegraphics[width=12cm]{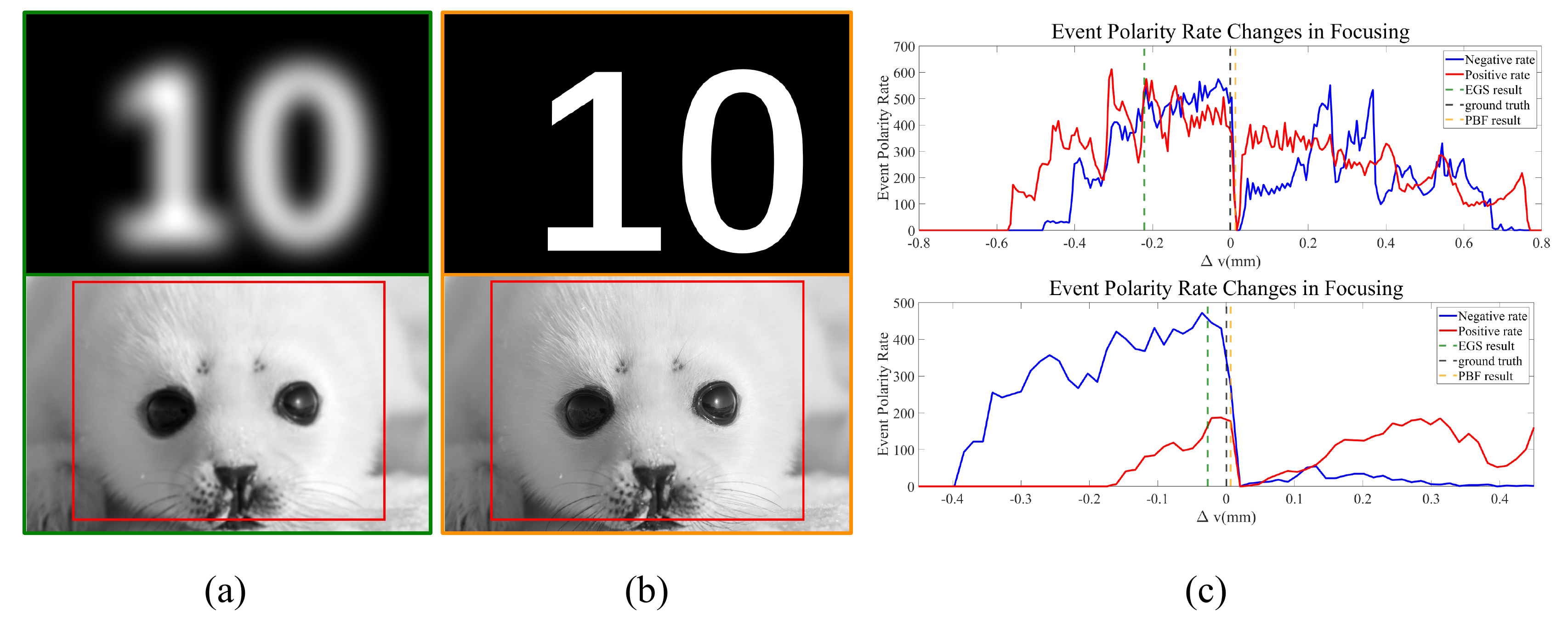}
\caption{Qualitative results of the simulated experiment. (a): Results of  EGS. (Top: $\Delta v$ = -221\si{\micro\meter}, 8 depths of focus. Bottom: $\Delta v$ = -27\si{\micro\meter}, 1 depth of focus).  (b): Results of PBF image (on focus). (c): EPR sequences. Dashed line: green for EGS result, yellow for PBF result, and black for ground truth. }
\label{fig: simulation}
\end{figure}

Two qualitative examples are shown in Fig.~\ref{fig: simulation}. The first row shows a sequence of focusing on a number with 100\% contrast. From the EPR curve in (c), the EGS algorithm finds the position in the sequence where the sum of PER and NER is greatest, marked by the green dashed line. Fig.~\ref{fig: simulation} (a) shows the corresponding image of the EGS result. On the contrary, the second row shows one of the cases where EGS performs relatively well. The red box represents the ROI for this case, ensuring that each feature inside it is sharp. In this case, the peaks of PER and NER sequences are not too far apart and are close to the ground truth position. Therefore, the event-rate-based method can also achieve an error within a depth of focus. In either example, the sequences of PER and NER are symmetrically distributed about the ground truth position and the PBF algorithm result (yellow dashed line in (c)) is extremely close to the ground truth and the corresponding images are sharp enough (shown in (b)), which proves the reliability of the PBF principle.

From the discussion of these two examples, it can be found that the distance separating the EPR peaks and the steepness of the peaks significantly affect the performance of the EGS algorithm in event camera focusing. In our simulation experiments, ER peaks are closer to the ground truth position for images with finer details, such as the second row of Fig.~\ref{fig: simulation}, while for images with clear boundaries and high contrast, such as the tree branches in Fig.~\ref{fig:branches example} and the figures in the first row of Fig.~\ref{fig: simulation}, the EPR peaks are farther away from the ground truth position. These intuitively demonstrate the shortcomings of the EGS algorithm. For the PBF algorithm, on the other hand, the symmetric distribution property of event polarities is universal, regardless of the texture and contrast of the scene selected.

The simulation experiments only consider events brought about by the change in the size of the convolution kernel during the focusing process, and can hardly mimic the noise in the real world. Therefore, in the next few subsections, more experiments are conducted on real event camera datasets to verify the robustness of our theory and algorithms.

\subsection{The Event-based AF Dataset (EAD)}
\label{subsection:42}
To better evaluate the performance of the PBF algorithm in the real-world event camera focusing process, experiments are conducted on public event camera AF dataset EAD~\cite{RN54}, comparing it with the state-of-art algorithm EGS~\cite{RN53}. 
\subsubsection{Experiment setup}
The EAD dataset uses the commercial event camera DAVIS 346 Color (resolution 346$\times$260) with a Fujifilm D17x7.5B-YN1 motorized zoom lens~\cite{RN54}, but the exact focal length of each sequence is unknown. The dataset includes different illumination levels (0.7 Lux to 23.35K lux), multiple real scenes (still and motion), and camera shake/still scenarios. 
This dataset is available in \cite{RN54}, and we also provide the numpy files of it in Dataset 1 (Ref.~\cite{EADnpy}). 

The hyperparameters used in the algorithm are as follows. EPR time interval is selected to 1 $\si{\milli\second}$. Regarding the EPR wavelet filter parameters, ``dmey'' is chosen as the base function for the 6-layer wavelet decomposition. Only the low-frequency component cA6 is reserved for wavelet reconstruction. The MSE investigation factor $k$ is selected to 0.5.

\subsubsection{Overall results for the EAD dataset}
Table~\ref{tab:addlabel} summarizes the performance between ER, EGS~\cite{RN53} and PBF algorithms in terms of mean absolute error (MAE) and root mean square error (RMSE) metrics.

\begin{table}[htbp]
  \centering
  \scriptsize
  \caption{Comparisons of ER, EGS, and PBF in EAD dataset. Our method outperforms in dark scenes, light static scenes and overall.}
    \begin{tabular}{c c c c c c c}
    \hline
    \multirow{2}[4]{*}{Method} & \multirow{2}[4]{*}{Metric} & \multicolumn{4}{c}{Scene Type} & \multirow{2}[4]{*}{Total}\\
            \cmidrule{3-6}
          &       &Light Static &Light Dynamic &Dark Static &Dark Dynamic &  \\
    \hline
    \multirow{2}[1]{*}{ER($\Delta t$ =0.055)} & MAE   & 98.6  & 105.3  & 59.3  & 87.4  & 87.6  \\
          & RMSE  & 112.8  & 119.2  & 73.2  & 113.9  & 106.4  \\
    \hline
    \multirow{2}[1]{*}{ER($\Delta t$ =0.065)} & MAE   & 91.4  & 74.3  & 58.0  & 67.1  & 72.7  \\
          & RMSE  & 107.3  & 87.6  & 71.9  & 103.7  & 93.7  \\
    \hline
    \multirow{2}[1]{*}{EGS} & MAE   & 77.3  & 29.0  & 54.1  & 64.9  & 56.3  \\
          & RMSE  & 98.9  & 33.4  & 71.7  & 96.7  & 79.7  \\
    \hline

    \multirow{2}[1]{*}{PBF} & MAE   & 47.8  & 64.0  & 54.3  & 39.0  & 52.5 \\
          & RMSE  & 51.8  & 87.3  & 65.1  & 52.2  & 65.7 \\
    \hline
    \end{tabular}%
  \label{tab:addlabel}%
\end{table}%

The proposed PBF algorithm outperforms the other event-based methods in the dataset as a whole and in all scenes except light dynamic scenes. The EGS algorithm performs better in the light dynamic scenes due to the fact that the focusing speed used in the EAD dataset is too slow compared to the camera motion, resulting in far more events generated by the motion than by focusing itself, which leads to a larger deviation in PBF. At the same time, EGS achieves better results because the optical flow of the images remains constant in these sequences, satisfying the assumption of the constant optical flow of the EGS algorithm.

It is worth noting that for focusing tasks, any focusing error of less than one depth of focus is acceptable. For the lens used in the EAD dataset, the depth of focus is about 50 \si{\micro\meter}. The evaluation of the algorithm robustness needs to focus more on the short board data, e.g., scenarios where the error exceeds 4-5 depths of focus. Compared to the EGS algorithm, which produces a maximum error of 236 µm (about 5 depths of focus) in the EAD dataset, our PBF algorithm produces an error of only 149 µm (about 3 depths of focus) in the worst case. 

In terms of running speed, the running time of the EGS algorithm on different sequences varies greatly (4\si{\milli\second} to 12s, average 2.6s), proportional to the number of events in the sequence. However, for our algorithm, the elapsed time is about 0.014s on every sequence. On average, our algorithm is 185 times faster than the speed of the EGS algorithm on the whole EAD dataset.

\begin{figure}[ht!]
\centering\includegraphics[width=10cm]{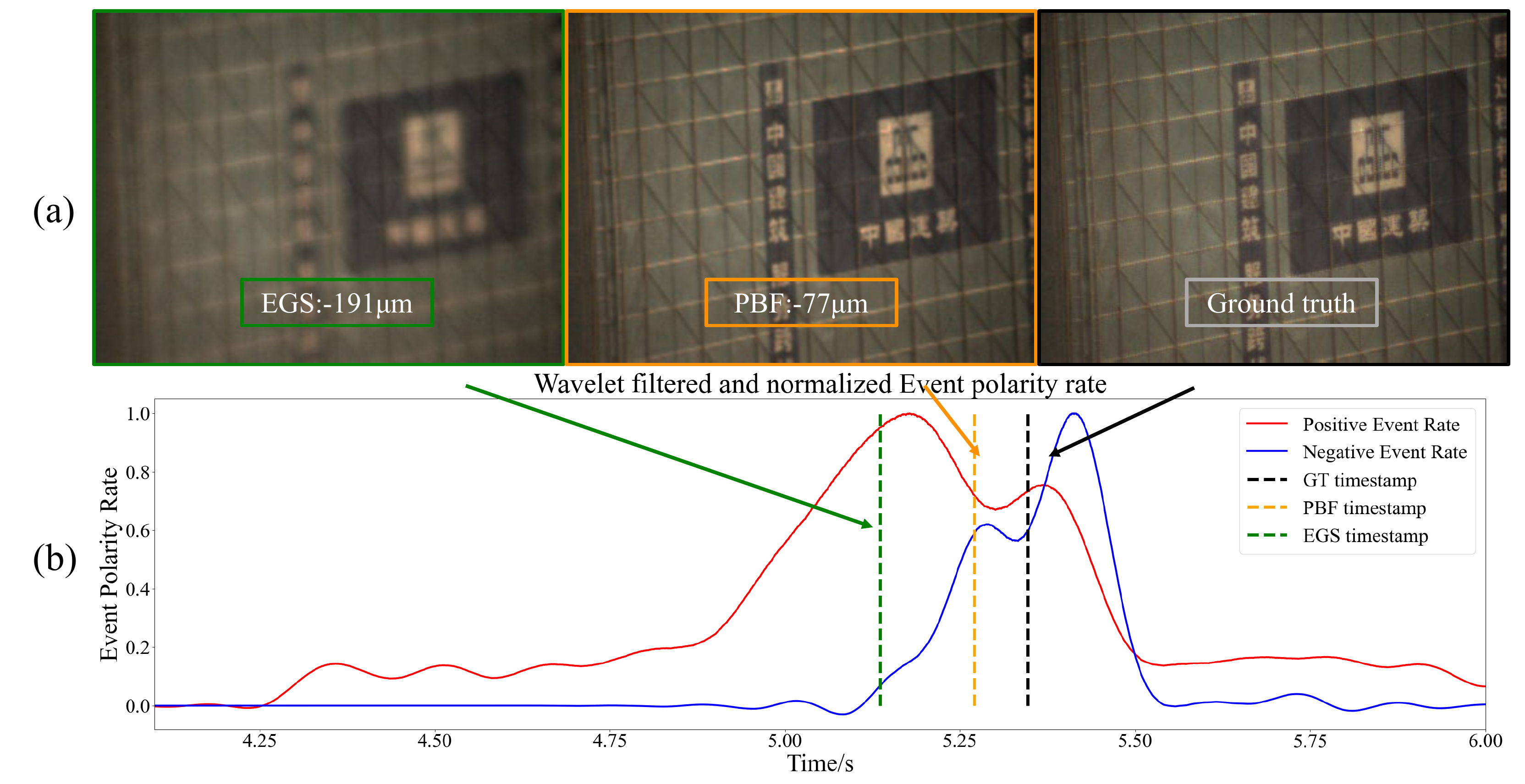}
\caption{Analysis on ``chinese light static'' sequence in EAD dataset. (a): APS frame images corresponding to the timestamps. (left to right: EGS result, PBF result, ground truth). (b): The symmetric EPR sequences after wavelet de-noise and polarity normalization.}
\label{fig: EAD_analysis}
\end{figure}

\subsubsection{Qualitative Analysis}
The sequences shown in Fig.~\ref{fig:related works} (a) are selected as an example to qualitatively compare the PBF algorithm with the EGS algorithm in the EAD dataset.

The ``chinese light static'' raw sequence of the EAD dataset shows a large difference between PER and NER, as shown in Fig.~\ref{fig:related works} (a). Fig.~\ref{fig: EAD_analysis} (b) shows the EPR sequences after wavelet transform and normalization mentioned in Algorithm~\ref{algorithm1}. Wavelet filtering eliminates the dense spikes in the raw sequences, making the variation of EPR with image distance $v$ changing more intuitive. The normalized PER and NER sequences show a clear symmetry with the ground truth timestamp. In this sequence, the PBF obtains a focus error of -77 \si{\micro\meter} (nearly 1 depth of focus) compared to the -191 \si{\micro\meter} focus error of the EGS (equivalent to about 4 depths of focus), as shown by the dashed lines in Fig.~\ref{fig: EAD_analysis} (b).

\subsection{Event camera high-speed AF dataset (EvHF)}
\label{subsection:43}
While the EAD dataset offers a variety of scenes for testing focusing algorithms, the image plane movement speed during its focusing process is only 1\si{\milli\meter}/\si{\second}. Such a slow focusing speed is not representative of real-world applications. Furthermore, the EAD dataset contains many scenarios where different parts of the scene have different focus positions, making it difficult to establish unambiguous ground truth. To evaluate the robustness of our PBF and EGS algorithms at a more realistic focusing motor speed of 10\si{\milli\meter}/\si{\second}, which is typical of current digital camera autofocus (AF) motors, our own Event camera high-speed AF (EvHF) dataset is collected.


\subsubsection{Experiment setup}

A DAVIS346 mono camera with a 35mm focal length, F1.4, C-Mount lens (depth of focus: 60 \si{\micro\meter}) is used for data acquisition. The camera is mounted on a ProScan III displacement stage from Prior Scientific, and the lens is fixed on the optical stage. The optical axis of the lens is adjusted to the center of the DAVIS sensor plane and perpendicular to it.
Due to the absence of screws to attach the C-Mount lens to the camera, there is a gap at the interface. To prevent stray light from entering the sensor surface through this gap, appropriate shading measures are implemented.

To determine the focus position, we initially move the displacement stage slowly and employ a traditional contrast-based focus function~\cite{RN50} to analyze the sharpness of the APS frame image obtained from the DAVIS camera. The focus position is recorded for future reference. During the formal focusing process, the displacement data from the stage is associated with the timestamp of the event camera by sending a high level TTL signal to the DAVIS346 whenever displacement data is acquired from the stage. To obtain a reasonable focus evaluation, it is necessary to ensure that all features inside ROI have the same depth. So an LCD placed perpendicular to the optical axis is chosen to focus on. The LCD screen is placed in an extremely dark scene to minimize the effect of external light reflecting off the surface.

By playing static or dynamic features on the screen and adjusting the brightness of the screen, 16 sets of focus data are obtained as shown in Table~\ref{tab:addlabel_2and3}. The raw data as well as numpy files is available in Dataset 2 (Ref.~\cite{highspeeddataset}).

The following explains the setup of dynamic scenes. The ``focusboard'' and ``panda'' sequences simulate irregular camera motion to mimic the effect of handheld shots. In the ``street'' sequence, scene motion is used to create a realistic portrayal of a moving vehicle. The most challenging sequence is the ``numbers'' sequence, which features scrolling numbers at 30 frames per second (fps). This sequence shows a significant challenge due to the high-speed numbers variation, which causes numerous noisy events output.

The hyperparameters used for the experiments on the EvHF dataset are the same as those used for the EAD dataset, shown in Section~\ref{subsection:42}.

\begin{table}[htbp]
  \centering
  \scriptsize
  \caption{Comparisons of PBF, EGS, and the ablation in the EvHF dataset.}
    \begin{tabular}{c c c c c c}
    \hline
    \multirow{1}[4]{*}{Scene} & \multirow{1}[4]{*}{Sequence name} & \multirow{1}[4]{*}{PBF error/\si{\micro\meter}} & \multirow{1}[4]{*}{EGS error/\si{\micro\meter}} & \multicolumn{2}{c}{Ablation error/\si{\micro\meter}} \\
        \cmidrule{5-6}
         &       &       &       & without filtering & without MSE \\
    \hline
    \multirow{3}[5]{*}{Dark Dynamic} & D\_D\_focusboard & 53    & -62   & 73    & -107 \\
          & D\_D\_numbers & 22    & 613   & -643  & 572 \\
          & D\_D\_panda & 13    & -28   & 13    & -47 \\
          & D\_D\_street & 41    & -61   & 101   & -99 \\
    \hline
    \multirow{3}[5]{*}{Light Dynamic} & L\_D\_focusboard & 58    & -212  & 73    & -197 \\
          & L\_D\_numbers & 6     & 231   & 6     & 216 \\
          & L\_D\_panda & 57    & -26   & 67    & -53 \\
          & L\_D\_street & 26    & -53   & 21    & -109 \\
    \hline
    \multirow{3}[5]{*}{Dark Static} & D\_S\_focusboard & 53    & -221  & 93    & -57 \\
          & D\_S\_numbers & 21    &599   & 61    & 571 \\
          & D\_S\_panda & 26    & 27    & 46    & -14 \\
          & D\_S\_street & 31    & -59   & 61    & -79 \\
    \hline
    \multirow{3}[5]{*}{Light Static} & L\_S\_focusboard & 58    & -192  & 93    & -107 \\
          & L\_S\_numbers & 16    & 188   & -4    & 146 \\
          & L\_S\_panda & 38    & -148  & 38    & -152 \\
          & L\_S\_street & 21    & -75   & 21    & -99 \\
    \hline
    \multicolumn{2}{c}{MAE/\si{\micro\meter}} & 34 & 175 & 88 & 164 \\
    \hline
    \multicolumn{2}{c}{Average time/\si{\milli\second} } & 4 & 146 & 3 & 1 \\
    \hline
    \end{tabular}%
  \label{tab:addlabel_2and3}%
\end{table}%

\subsubsection{Overall results for the EvHF dataset}
The performance of both EGS algorithm and our proposed PBF algorithm are evaluated on EvHF dataset, and the detailed results are shown in Table~\ref{tab:addlabel_2and3}. In terms of focusing tasks, errors within one depth of focus (60 \si{\micro\meter}) are considered acceptable. The PBF algorithm has achieved an error within 1 depth of focus for all sequences in the dataset, while the EGS algorithm shows a worst error of 10 depths of focus (613 \si{\micro\meter}). On average, the PBF algorithm has an MAE of 34 \si{\micro\meter} on the EvHF dataset, while the EGS is 175 \si{\micro\meter}, which is a 5-fold improvement in accuracy in the high-speed scenario. In terms of speed, our algorithm is 36 times faster than the speed of the EGS algorithm on average, referring to the last row of Table~\ref{tab:addlabel_2and3}.

To provide a visual comparison of the effectiveness of the PBF algorithm and the EGS algorithm, some examples from the dataset are elaborated on in the next subsection.

\begin{figure}[ht!]
\centering\includegraphics[width=11cm]{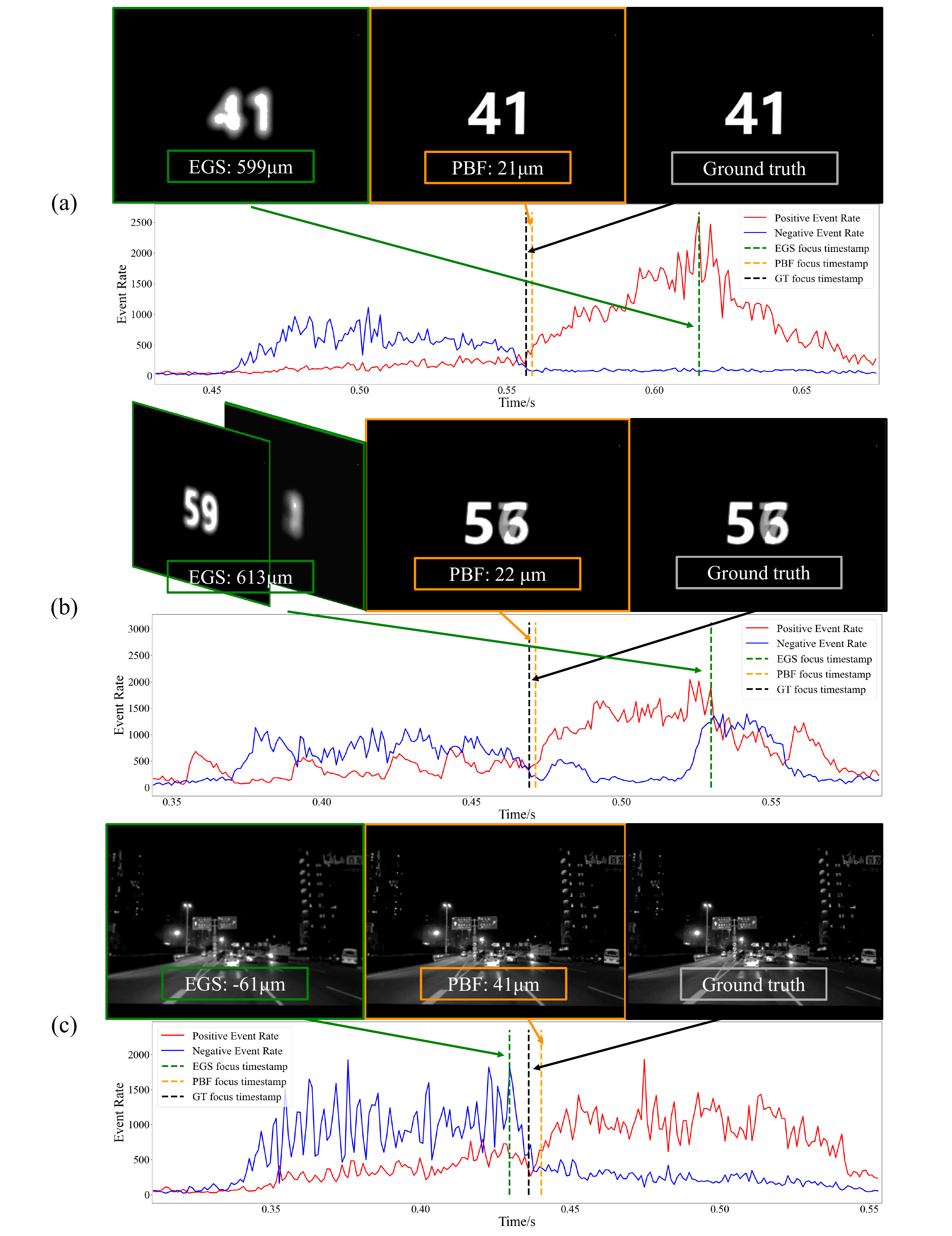}
\caption{Comparison of the qualitative results from PBF algorithm and EGS algorithm in EvHF dataset. (a): ``Numbers'' in dark static scene. (b): ``Numbers'' in dark dynamic scene. The jump of the number from ``59'' to ``1'' occurs around the EGS timestamp. (c): ``Street'' in dark dynamic scene. In each sub-figure, above: APS images corresponding to the timestamps, below: EPR plot. 
}\label{fig: High-speed result}
\end{figure}

\subsubsection{Qualitative Analysis}

Fig.~\ref{fig: High-speed result} (a) shows the EPR sequences when focusing a number '41' in a dark stationary environment. As can be seen, in the dark environment, there are generally more positive events than negative events due to the different trigger thresholds. Since the goal of the EGS algorithm is to find the position with the highest ER in the whole focusing sequence, it faithfully finds the peak position of the ER (i.e., the sum of PER and NER), which is 599 \si{\micro\meter} (equivalent to 10 depths of focus) away from the true focus position, resulting in an unacceptably blurry picture. The PBF algorithm, based on the symmetric relationship of PER and NER, obtains a focusing error (21 \si{\micro\meter}) of less than one depth of focus (60 \si{\micro\meter}). At the same time, due to wavelet denoising, the high-frequency noise prevalent in the EPR sequence does not interfere with the determination of the center of symmetry of the sequence. 

When examining the dataset ``numbers'' in the dark dynamic scene shown in Fig.~\ref{fig: High-speed result} (b), the PBF algorithm gives fairly stable results, while the EGS algorithm gives an error of nearly 10 depths of focus (613 \si{\micro\meter}). From the EPR graph, there are peaks of PER and NER near the green EGS timestamp. According to the APS images, such peaks come from the jump of the number from ``59'' to ``1''. Such a large change in brightness brings more than the number of events near the focus position, so the EGS algorithm based on the ER evaluation fails in this case. The PBF algorithm, on the other hand, maintains robustness to such sudden event surges by exploiting the symmetric nature of the PER and NER in the focusing process. In the dark high-speed focus scenario, the APS sensor of DAVIS is able to capture only 8 images at a frame rate of about 40 fps for the focus motion completed in 0.2 s, which is not sufficient for the traditional contrast-based method of focus evaluation. And, due to the high-speed motion of the scene, image trailing occurs during the APS exposure time of about 10 $\si{\milli\second}$, as in the image pointing to GT timestamp shown in Fig.~\ref{fig: High-speed result} (b), which also means that it is not feasible to use the traditional image focus evaluation method.

The EGS algorithm obtains a relatively good result in the scenario shown in Fig.~\ref{fig: High-speed result} (c). As can be seen from the EPR graph, the peak position of the sum of PER and NER in this scenario is close to the GT position (-61 \si{\micro\meter}), thus barely meeting the requirements. The PBF algorithm still faithfully finds the center of symmetry position of the PER and NER curves, and thus the error with the GT position is less than one depth of focus. From the simulation experiments in Section~\ref{subsection:41}, it can be learned that the distance between the PER and NER peaks separation is related to the texture and contrast of the scene.

\subsubsection{Algorithm speed comparison}
In high-speed focusing scenarios, the speed of the algorithm is as important an evaluation indicator as the accuracy. The computational complexity of EGS algorithm $O(N_e)$, which means that the algorithm time consumption is linear to the number of events. In addition, the actual program runtime is consumed heavily in cache reads and writes because of the need to repeatedly access tens of millions of event data. 
In contrast, the original data used by the PBF algorithm is two integer arrays with a length of hundreds or thousands. Such arrays of EPR are available synchronously during the event data acquisition. The running time of the PBF algorithm is mainly spent on the EPR filtering and PBF MSE calculation, which can be done in a few milliseconds even on a single-threaded CPU.

Taking the sequence shown in Fig.~\ref{fig: High-speed result} (a) as an example, the total number of events is 885,024, the length of the EPR sequence is 243, the running time of the Matlab code of the EGS algorithm is 293 \si{\milli\second}, while the running time of the python code of the PBF algorithm is 4 \si{\milli\second}. In the case of a higher number of events, the EGS algorithm time consumption increases linearly with the number of events, while the PBF algorithm time consumption is almost constant.
\subsection{Robustness to brightness changes during focusing}
\label{subsection:brightness change}
The sensitivity of event cameras to brightness changes makes them particularly susceptible to influences such as stroboscopic light sources and sudden brightness changes. Such sudden changes in brightness are common, such as the widespread use of 50Hz stroboscopic light sources at night and burst of flashes from surrounding areas during nighttime photography. Experiments are conducted on the two brightness change scenarios during focusing, to verify the robustness of our algorithm. The rest of the experimental parameter settings and algorithmic hyperparameter settings remain the same as in section \ref{subsection:43}. The depth of focus of the lens remains at 60 \si{\micro\meter}.
\subsubsection{Robustness to stroboscopic light source}
\begin{figure}[htbp]
    \centering
    \vspace{-5pt}
    \includegraphics[width=11cm]{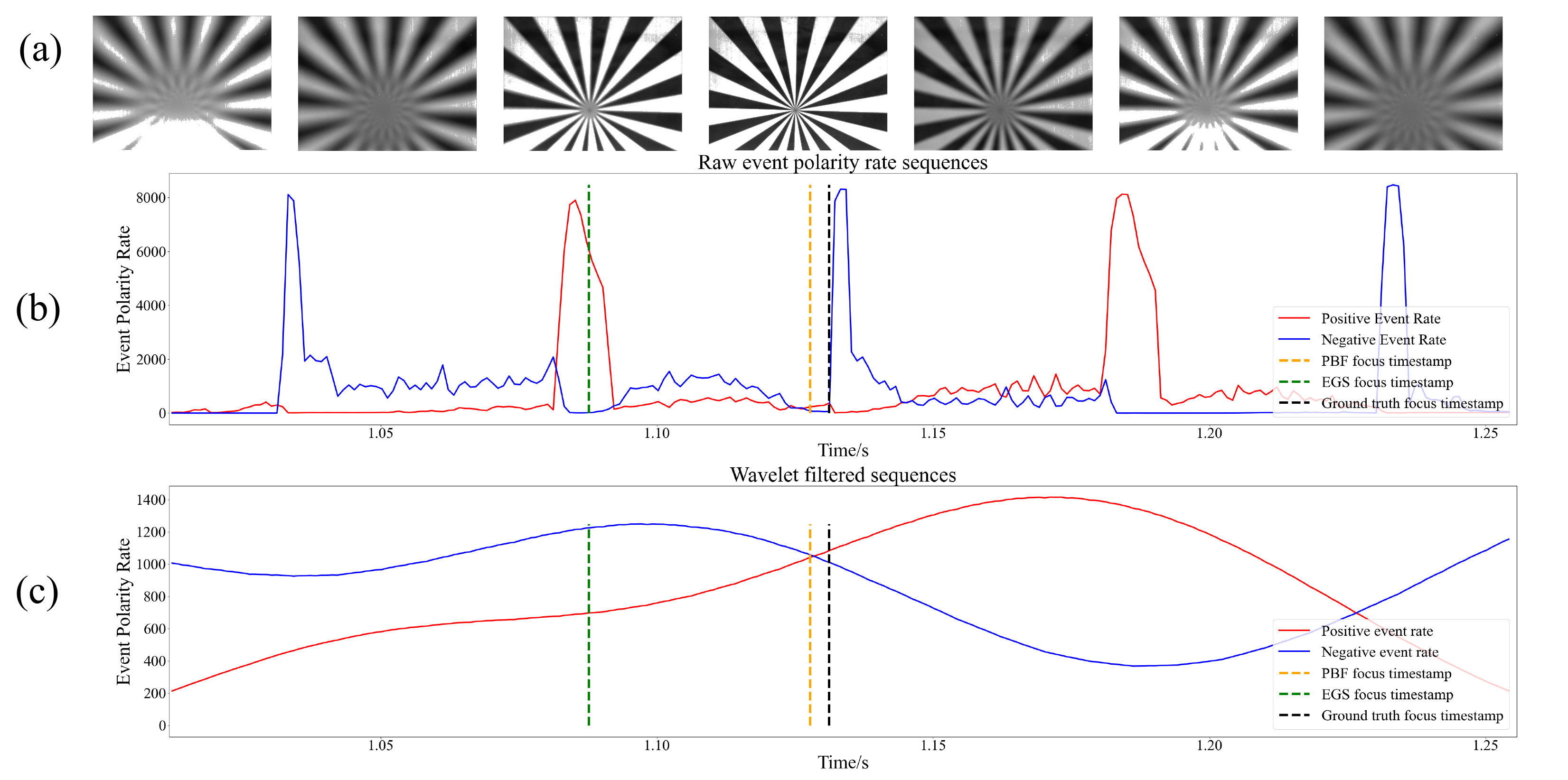}
    \vspace{-7pt}
    \caption{Experiment results when the event camera is affected by intense stroboscopic light sources during the focusing process. (a) APS frame images output synchronously with event signals. (b) Raw event polarity rate sequences. (c) The filtered sequences of event polarity rate.}\label{fig:stroboscopic_light_source}
\end{figure}

In the experiment, the focusboard image is placed in front of a 20Hz strobe flashlight and brightness keeps changing throughout the focusing process. From the APS frame images synchronously outputted with event signals from Fig.~\ref{fig:stroboscopic_light_source} (a), it can be intuitively seen that there is a difference between flashing light and dark. At the same time, the raw event polarity rate sequences in Fig.~\ref{fig:stroboscopic_light_source} (b) also confirm the aforementioned phenomenon. Due to the excessive generation of noise events compared to the motion caused by focusing, the event rate-based EGS algorithm fails in this scenario (error: -435 \si{\micro\meter}, 7 depths of focus) as it assumes that the position with the highest event rate corresponds to the focus position. This failure can be attributed to the significant impact of strobe brightness changes, which leads to a distorted event rate distribution. However, our PBF algorithm based on the event polarity rate remains robust in this scenario (error:-35 \si{\micro\meter}, less than 1 depth of focus). As shown in Fig.~\ref{fig:stroboscopic_light_source} (c), the wavelet transform preserves the  slow-varying signal related to the focusing motion, while effectively filtering out the 20Hz stroboscopic noise signal with a signal amplitude close to four times.

\subsubsection{Robustness to arbitrary brightness changes}
\begin{figure}[htbp]
    \centering
    \vspace{-5pt}
    \includegraphics[width=11cm]{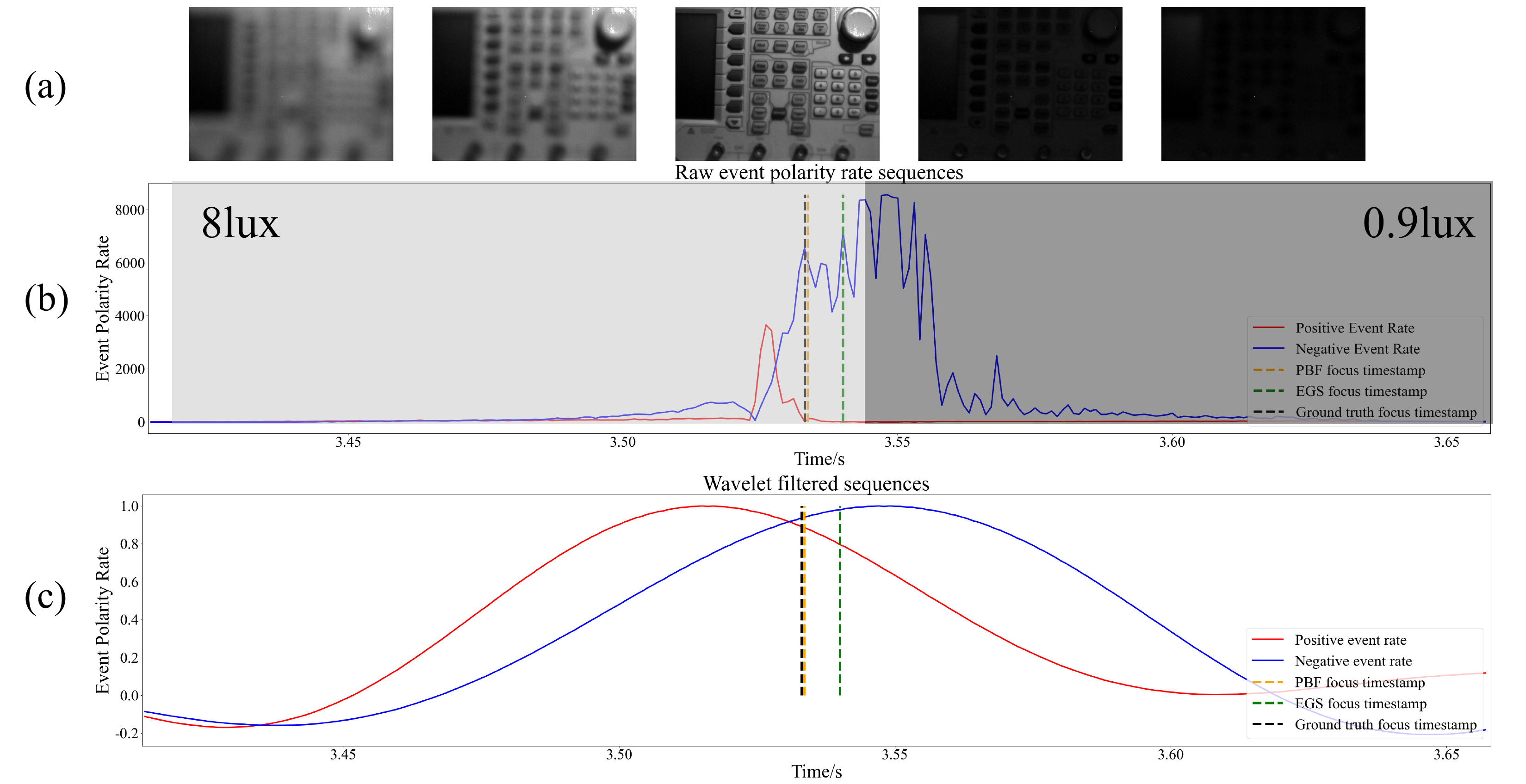}
    \vspace{-7pt}
    \caption{Experiments of exploring the robustness to arbitrary brightness change. (a) APS frame images output synchronously with event signals. (b) Raw event polarity rate sequence. (c) The filtered sequence of event polarity rates.}\label{fig:arbitrary_brightness_change}
\end{figure}
The experiment conducted in this section involves a scene where the surface brightness of the object to be focused undergoes arbitrary changes under relatively low illumination conditions. The robustness of our algorithm is explored with the experiment settings.

In the experiment, a flashlight is used to illuminate the surface of the object of interest, specifically an oscilloscope, with an initial surface illumination of approximately 8 lux. At an arbitrary point in the focusing process, the flashlight is turned off, resulting in a significant drop in surface illumination to approximately 0.9 lux,  as illustrated in Fig.\ref{fig:arbitrary_brightness_change} (a). 
The significant illumination change results in a substantial increase in negative events captured by the event camera, as depicted in Fig.~\ref{fig:arbitrary_brightness_change} (b). This phenomenon disrupts the luminance invariance assumption underlying the EGS algorithm, leading to a focus position that is noticeably influenced by the time point of the abrupt luminance change (error: 179 \si{\micro\meter}, 3 depths of focus). Regarding our PBF algorithm, due to the suppression of time-varying random noise by the wavelet filtering, in this scenario, as Fig.\ref{fig:arbitrary_brightness_change}(c) demonstrates, the filtered event polarity rate sequence successfully preserves the slow-varying signal related to the focusing motion, resulting in an error of 5 \si{\micro\meter} (less than 1 depth of focus).

\subsection{Ablation study.}
\label{subsec:ablation}
The proposed PBF algorithm consists of two important parts: the filtering based on wavelet transform, and the focus position determination based on the PBF MSE evaluation function. Ablation experiments are conducted on these two parts separately, and the overall results are shown in the last two columns of Table~\ref{tab:addlabel_2and3}. All the ablation experiments are conducted on our EvHF dataset in Section~\ref{subsection:43}. In the first part of the ablation experiment, the wavelet filter module is removed and the original un-denoised EPR sequences are used for MSE computation. In the second part of the ablation experiment, the wavelet filter module is kept but the ER algorithm is used to make a focus evaluation.


\subsubsection{Quantitative results}
The last column of Table~\ref{tab:addlabel_2and3} shows the quantitative result of ablation experiments. The ablation of the filtering step only significantly affects some sequences with a low signal-to-noise ratio, (e.g., ``numbers'' in dark dynamic scene), while the others still perform well. However, when ablating the core PBF MSE evaluation module, poor results are obtained in most cases.

\subsubsection{Qualitative Analysis}
The ``numbers'' sequence in the dark dynamic scene shown in Fig.~\ref{fig: High-speed result} (b) is selected to analyze the results of the ablation experiment. 

\begin{figure}[ht!]
    \centering\includegraphics[width=11cm]{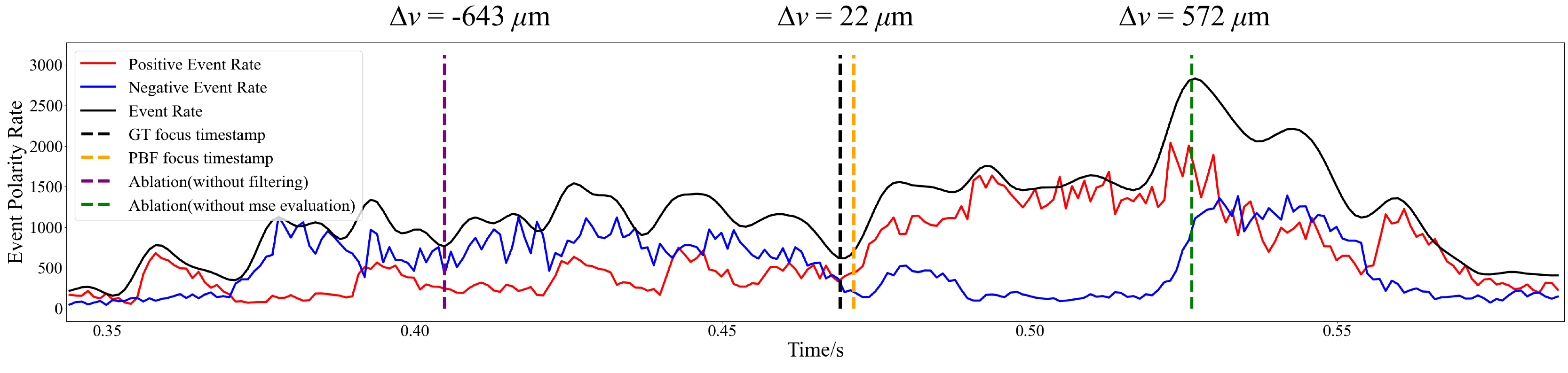}
    \caption{Results of the ablation experiment (data sequence ``numbers'' in dark dynamic scene). The dashed line: black for the ground truth, orange for PBF result, purple for the result ablating filtering, and green for the result ablating MSE evaluation.}\label{fig:ablation}
\end{figure}

The wavelet filtering de-noise process plays an important role in improving the robustness. After the ablation of the de-noise process, the events brought by the high-speed changing numbers generate noise with a frequency of 30Hz in the EPR sequence, as shown in Fig.~\ref{fig:ablation}. Without the filtering operation, the MSE evaluation function of the PBF algorithm is no longer a single valley function and therefore generates unpredictable errors in finding the symmetry center. Near the 
purple dashed line, the positive and negative events caused by the scene motion also show some symmetry on the time axis because of the 30Hz scene movement, and the MSE evaluation function reaches a global minimum at this position, thus producing erroneous results.

The PBF MSE evaluation function is the core part of our algorithm. When this part is ablated, even if the noise in the ER sequence is filtered using the wavelet transform, the results obtained using the ER maximum-based method do not differ much from those obtained by the EGS algorithm, referring to Table~\ref{tab:addlabel_2and3}. The reason for such a result is the same as the EGS algorithm. In fact, the EGS algorithm's dichotomous lookup method also plays the same function as wavelet transform de-noising.

\section{Conclusion}
In this paper, we (1) explain the mechanism of events output in the AF process, (2) find the relationship between the EPR sequence and the focus position, and (3) derive the Polarity-based auto-focus (PBF) algorithm. Our proposed algorithm is verified on the event camera AF public EAD dataset and compare it compared with the state-of-art event camera AF algorithm event-based golden search (EGS). Our PBF algorithm shows higher accuracy and faster speed. We also collect our event camera high-speed focusing (EvHF) dataset featuring scenes with
varying lighting conditions and motion states. The
supplement experiments on our dataset verify the robustness and real-time performance of our proposed PBF algorithm in high-speed focus scenarios. The experimental results provide compelling evidence that the proposed PBF algorithm can ensure highly accurate focusing performance, achieving errors below 1 depth of focus across diverse scenes with varying brightness and motion states.

\section{Discussion}
Currently, due to the limited spatial resolution of the event camera device, the experiment employs event information from all pixels. In theory, higher resolution would allow for selecting regions of interest (ROIs) at different positions in the image plane for separate focus evaluation. Utilizing knowledge of the lens parameters and image distance, the distance from the object side to the lens can be determined via the object-image relationship, enabling three-dimensional spatial recovery based on the focus variation method. However, when examining a smaller ROI, such as a single pixel, the effect of feature shift due to breathing must be taken into consideration, as discussed in Section~\ref{subsubsection:323}.

\begin{backmatter}
\bmsection{Funding} This work is supported by the National Key R\&D Program of China, under Grant No. 2022YFF0705500, National Key R\&D Program of China (2022YFB3206000).


\bigskip
\bmsection{Disclosures}
The authors declare no conflicts of interest.

\bigskip
\bmsection{Data availability} Data underlying the results presented in this paper are available 
in Ref.~\cite{PBFcode,EADnpy,highspeeddataset,simulationcode}.

\end{backmatter}


\bibliography{PBF}

\end{document}